\pdfoutput=1

\documentclass[11pt]{article}

\usepackage[]{EMNLP2022}

\usepackage{times}
\usepackage{latexsym}

\usepackage[T1]{fontenc}

\usepackage[utf8]{inputenc}

\usepackage{microtype}

\usepackage{inconsolata}
\usepackage{times}
\usepackage{latexsym}
\usepackage{bm}
\usepackage{amsthm,amsmath,amssymb}
\usepackage{mathrsfs}
\usepackage{epsfig}
\usepackage{graphicx}
\usepackage{booktabs}
\usepackage{verbatim}
\usepackage{enumitem,kantlipsum}
\usepackage{tipa}
\usepackage{algorithm}
\usepackage{algorithmicx}
\usepackage{algpseudocode}

\newcommand{\seq}[1]{\boldsymbol{#1}}
\usepackage{color} 
\definecolor{mypink2}{RGB}{219, 48, 122}
\definecolor{orange}{RGB}{255, 147, 00}
\definecolor{jrcolor}{RGB}{100, 150, 225}
\definecolor{jrcomment}{RGB}{70, 200, 150}
\definecolor{grey}{RGB}{166, 166, 166}

\title{$N$-gram Is Back: Residual Learning of Neural Text Generation \\ with $n$-gram Language Model}

\author{Huayang Li$^{\clubsuit}$~~~~Deng Cai$^{^\heartsuit}$~~~~Jin Xu$^\diamondsuit$~~~~Taro Watanabe$^{\clubsuit}$\\
  $^\clubsuit$Nara Institute of Science and Technology~~~~
  $^\heartsuit$The Chinese University of Hong Kong~~~\\ $^\diamondsuit$Institute for Interdisciplinary Information Sciences, Tsinghua University\\
  \texttt{\{li.huayang.lh6, taro\}@is.naist.jp}~~~ \texttt{thisisjcykcd@gmail.com} \\ \texttt{xujin21@mails.tsinghua.edu.cn}\\}

\begin{document}
\maketitle
\begin{abstract}
$N$-gram language models (LM) have been largely superseded by neural LMs as the latter exhibits better performance. However, we find that $n$-gram models can achieve satisfactory performance on a large proportion of testing cases, indicating they have already captured abundant knowledge of the language with relatively low computational cost. With this observation, we propose to learn a neural LM that fits the residual between an $n$-gram LM and the real-data distribution. The combination of $n$-gram and neural LMs not only allows the neural part to focus on the deeper understanding of language but also provides a flexible way to customize an LM by switching the underlying $n$-gram model without changing the neural model. Experimental results on three typical language tasks (i.e., language modeling, machine translation, and summarization) demonstrate that our approach attains additional performance gains over popular standalone neural models consistently. We also show that our approach allows for effective domain adaptation by simply switching to a domain-specific $n$-gram model, without any extra training. Our code is released at \url{https://github.com/ghrua/NgramRes}.
\end{abstract}

\section{Introduction}
$N$-gram language model (LM) was widely adopted in a broad range of natural language processing (NLP) applications, such as input method \cite{chen-etal-2019-federated}, statistical machine translation \cite{brown1990statistical}, and audio speech recognition \cite{bahl1983maximum}. However, with the development of deep learning, neural LMs have gradually taken the place of $n$-gram LMs and became the new standard in recent literature \cite{merity2016pointer, vaswani2017attention, radford2019language}. One critical reason is the superior performance of neural LMs, e.g., the GPT-2 model \cite{radford2019language} can generate text near the human level, outperforming $n$-gram LMs by large margins.

\begin{figure}[t]
    \begin{center}
        \includegraphics[width=0.9\linewidth]{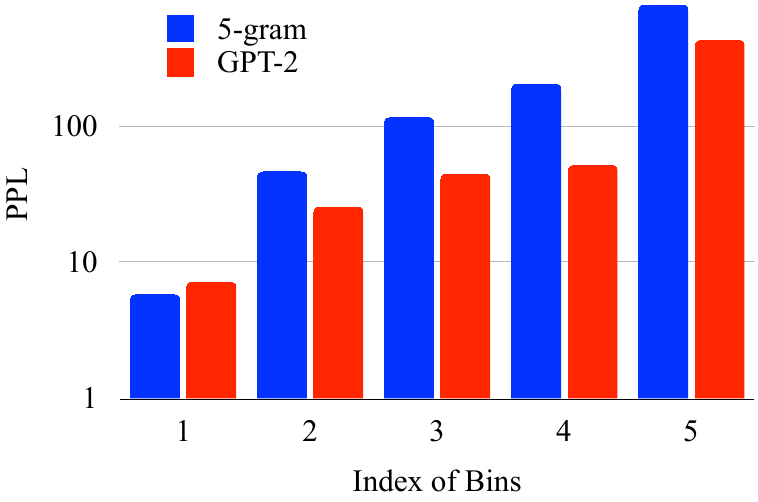}
    \end{center}
        \caption{Sentence-level perplexity (PPL) of $5$-gram LM and GPT-2 LM on the validation dataset of wikitext-103. We sort sentences in the validation dataset according to their $5$-gram PPL scores, and collect them into 5 bins with an equal number of sentences. The reported PPL score of each bin is the average over the sentences in it, and the y-axis uses a logarithmic scale. Details of the dataset and LMs are shown in section \ref{sec:exp_lm}. \label{fig:illustration}}
\end{figure}

Despite that neural LMs have surpassed $n$-gram models at the macro level, we find that $n$-gram LMs are still attractive: they are able to achieve satisfactory performance on a large proportion of testing cases at a much lower cost than neural LMs. As observed in Figure \ref{fig:illustration}, our preliminary experiments show that the performance of $5$-gram LM is close to the \textsc{GPT-2} model trained from scratch on 3 out of 5 bins ( 1, 2, and 5). Moreover, the performance of $5$-gram on the first bin is slightly better than \textsc{GPT-2}. Because training a neural LM is much more expensive, spending effort on learning the knowledge that can be cheaply captured by $n$-gram seems a waste.

Inspired by the above observation, we propose to learn a neural LM that focuses on the information gap that has not been captured by an $n$-gram model: $\mathcal{F}:= \mathcal{G} - \mathcal{Q}$, where $\mathcal{G}$ and $\mathcal{Q}$ are the real-data distribution and the $n$-gram prediction distribution respectively, which is in a similar spirit to residual learning \cite{he2016deep}. More concretely, we combine the logits (the unnormalized probability scores before \texttt{softmax} layer) of a neural model and those derived from an $n$-gram model. The joint neuro-symbolic system at least brings two appealing characteristics. First, since the neural model stands on the shoulders of the shallow $n$-gram LM, it can concentrate on deeper understanding. Second, the underlying $n$-gram LM can be purposefully switched without changing the neural model, which offers great flexibility in tackling scenarios such as domain adaptation. That is, we can adapt the model to a specific domain by changing the underlying $n$-gram LM in a plug-and-play manner, without changing any parameters of the neural model.

We conduct extensive experiments to evaluate the proposed approach. Experiments on the standard benchmarks of three typical language tasks, including language modeling, machine translation, and summarization, show that our approach can improve the performance of recent state-of-the-art neural models consistently and considerably. For example, our approach outperforms popular baseline models by at least 0.7 PPL scores on the wikitext-103 dataset for language modeling, 0.65 BLEU scores on average on IWSLT datasets for machine translation, and 0.36 ROUGE-L scores on the CNN/DailyMail dataset for summarization. Moreover, on the language modeling task, when switching the underlying $n$-gram LM to a particular domain-specific one (e.g., IT, Koran, Law, Medical, and Subtitles) in a plug-and-play manner, our model can reduce the PPL by 5.4 points on average without any domain-specific training of the neural part. Remarkably, the performance of our approach is even close to fine-tuning the whole model on domain-specific corpora.

Our contributions are three-fold:
\begin{itemize}
    \item We propose a residual learning approach for two heterogeneous structures, i.e., $n$-gram and neural LMs, which forces the neural LM to approximate the information gap that has not been captured by $n$-gram LM.
    \item Our approach is able to improve the performance of recent state-of-the-art neural models consistently and considerably on language modeling, machine translation, and summarization.
    \item  Experiments on domain adaptation demonstrate that our approach can effectively and cheaply adapt the model to a specific domain by changing the used $n$-gram LM in a plug-and-play manner, without changing any parameters of the neural model. 
\end{itemize}

\section{Related Work}

\paragraph{Language Model} The $n$-gram language model (LM) has been widely used in lots of applications of natural language processing (NLP) since a long time ago \cite{jurafsky2000speech}. The emergence of advanced smoothing technologies makes the $n$-gram model able to provide a better estimation of human languages \cite{kneser1995improved, chen1996empirical, heafield-etal-2013-scalable}. In statistical machine translation \cite{brown1990statistical} and automatic speech recognition \cite{bahl1983maximum}, the decoder-side $n$-gram model is critical to estimate the quality of generated candidates. In recent literature on input methods, the $n$-gram LM is still the most popular choice for providing word suggestions \cite{huang2015input,chen-etal-2019-federated}, because of its low cost and low latency.

However, with the development of deep neural networks, the macro-level performance of neural LM has surpassed that of $n$-gram LM by a large margin. Comparing with the $n$-gram LM, one big advantage of the neural LM basing on recurrent neural network \cite{hochreiter1997long, chung2014emperical} and attention neural network \cite{vaswani2017attention, radford2019language} is their ability to modeling long-distance dependencies \cite{grave2016improving}. The success of neural LM can also be observed in the big improvement achieved in lots of downstream tasks, e.g., text generation \cite{DBLP:conf/iclr/HoltzmanBDFC20,DBLP:conf/iclr/WelleckKRDCW20, DBLP:journals/corr/abs-2202-06417, DBLP:journals/corr/abs-2206-02369, li2022survey, cai2022recent}, machine translation \cite{bahdanau2015neural, Luong-Manning:iwslt15, vaswani2017attention,cai2021neural} and summarization \cite{ li-etal-2017-deep, see2017get, bi2020palm}. 

Although neural LM has outperformed $n$-gram LM at the macro level, we find that $n$-gram LM can achieve satisfactory performance on a large portion of testing cases. Since the training cost of neural LM is much more expensive and the model capacity is fixed, we hypothesize that it is not necessary to train the neural LM to learn the knowledge that can be captured by $n$-gram LM at a much lower cost. Therefore, we propose a residual learning method to let the neural LM learn the gap of knowledge that has not been captured by $n$-gram LM.

\paragraph{Residual Learning} Residual learning is a useful technique for lots of neural networks in computer vision (CV) and natural language processing (NLP). \citet{he2016deep} propose deep residual learning to alleviate the training difficulties of deep models, which has been the backbone of lots of tasks in CV. In NLP, \citet{wang2016recurrent} and \citet{prakash2016neural} use the residual learning technique to train deep recurrent neural networks for text generation. Different from previous works that conduct residual learning over different layers, \citet{werlen2018self} propose to aggregate the information of historical predictions using residual learning. In \citet{he2021realFormer}, they use the residual learning to propagate attention scores across different layers of the Transformer-based model.

Most of these works conduct residual learning over homogeneous model structures, e.g., stacked identical layers of the same model. In our work, we use residual learning to combine the neural and symbolic models, i.e., learn a neural LM that approximates the information that has not been captured by the $n$-gram model.

\section{Background}

Models that estimate the probabilities of sequences of words are called language models (LM) \cite{jurafsky2000speech}. Let $\seq{x} = \{x_1, x_2, ..., x_L\}$ be a sequence of words with length $L$. The probability of $P(\seq{x})$ can be formalized according to the chain rule of probability:

\begin{align}
    P(\seq{x}) &= P(x_1)P(x_2|x_1)\dots P(x_L|\seq{x}_{1}^{L-1}) \nonumber\\
               &= \prod_{k=1}^L P(x_k|\seq{x}_{1}^{k-1}),
\label{eq:prob}\end{align}
where $\seq{x}_{1}^{k-1}$ is called the prefix or context of $x_k$. In this section we will briefly introduce two kinds of language models, the $n$-gram and neural language models, to compute the probability in Eq. (\ref{eq:prob}).

\subsection{$N$-gram Language Model}
Among lots of variants of $n$-gram LMs, the $n$-gram LM with modified Kneser-Ney smoothing is widely adopted in lots of related tasks, because of its low perplexity and efficiency \cite{kneser1995improved, chen1996empirical, heafield-etal-2013-scalable}. Like most $n$-gram LMs, the Kneser-Ney approximates the entire context $x_{1}^{k-1}$ in Eq. (\ref{eq:prob}) by the last $n-1$ words in the context:
\begin{align}
    P(x_k|\seq{x}_{1}^{k-1}) &\approx P_{NG}(x_k | \seq{x}_{k-n+1}^{k-1}).
\label{eq:approx}\end{align}
In Kneser-Ney algorithm, the estimation of $P_{NG}(x_k | \seq{x}_{k-n+1}^{k-1})$ is defined according to a recursive equation:
\begin{align}
    P_{NG}(x_k|\seq{x}_{k-n+1}^{k-1}&) = U(x_k|\seq{x}_{k-n+1}^{k-1}) + \nonumber\\
    &\quad\quad b(\seq{x}_{k-n+1}^{k-1}) P_{NG}(x_k|\seq{x}_{k-n+2}^{k-1}), \label{eq:kn} \\
U(x_k|\seq{x}_{k-n+1}^{k-1}) &= \frac{c(\seq{x}_{k-n+1}^k) - d}{\sum_wc(\seq{x}_{k-n+1}^{k-1}w)}, \nonumber
\end{align}
where $w$ indicates a word appears after $\seq{x}_{k-n+1}^{k-1}$, $b(\cdot)$ is the backoff value for lower-order estimation, $c(\cdot)$ is the adjusted counts, $d$ is the discounts for smoothing \cite{jurafsky2000speech, heafield-etal-2013-scalable}\footnote{More details about adjusting counts and computing the backoff values and discounts are shown in \citet{jurafsky2000speech} and \citet{heafield-etal-2013-scalable}.}. According to Eq. (\ref{eq:kn}), Kneser-Ney allows us to assign probabilities for unseen $n$-grams (e.g., 5-grams), using the lower-order information (e.g., 4-, 3-, or even uni-grams).

\subsection{Neural Language Model}
An neural LM typically estimates the probability of $x_k$ based on the whole context $\seq{x}_{1}^{k-1}$. The parameter $\theta$ of a neural LM is optimized through the following MLE loss:
\begin{align}
\mathcal{L}_{NU} &= \sum_{\seq{x} \in \mathcal{D}}\sum_{k=1}^{L} \log P_{NU}(x_k|\seq{x}_{1}^{k-1}; \theta)
\label{eq:mle}\end{align}
where $\mathcal{D}$ is the training dataset. The probability of $P_{NU}(x_k| \cdot)$ is computed by:
\begin{align}
    P_{NU}(x_k|\seq{x}_{1}^{k-1}; \theta) = \mathrm{softmax}(\phi(\seq{h}_k))[x_k],
\label{eq:neural_prob}\end{align}
where $\seq{h}_k$ is the hidden vector output by the last layer of an neural LM, e.g., the GPT-2 model \cite{radford2019language} or LSTM model \cite{grave2016improving}. The $[x_k]$ is defined as taking the component regarding to $x_k$ in a vector, i.e., the probabilistic distribution got from $\mathrm{softmax}$ in this equation. The $\phi(\cdot)$ is a linear layer that transforms the hidden vector $\seq{h}_k$ to a vector in the vocabulary space, which is also called the logits. 

\section{Methodology\label{sec:approach}}

\subsection{Motivation}
The main idea of our work is to use the neural LM to approximate a residual function.
Given the context $\seq{x}_{1}^{k-1}$ in the language modeling task, let us consider $\mathcal{G}(\seq{x}_{1}^{k-1})$ as the golden-truth distribution of the next word, and
\begin{equation}
    \mathcal{Q}(\seq{x}_{1}^{k-1}) = P_{NG}(X|\seq{x}_{k-n+1}^{k-1})\label{eq:ngram_dist}
\end{equation}
as the prediction distribution of the $n$-gram LM, where $X$ is the random variable and the probability $P_{NG}(X=x_k|\seq{x}_{k-n+1}^{k-1})$ is calculated according to Eq.~(\ref{eq:kn}). Since the $n$-gram distribution $\mathcal{Q}(\seq{x}_{1}^{k-1})$ has captured abundant information of the language as we discussed in the introduction, one interesting question is: can we use a neural LM to approximate the residual function $\mathcal{F}(\seq{x}_{1}^{k-1}) := \mathcal{G}(\seq{x}_{1}^{k-1}) - \mathcal{Q}(\seq{x}_{1}^{k-1})$? This is similar to the residual learning in \citet{he2016deep}. If it is possible, we can release the burden of neural LMs on learning the information that has been captured by $n$-gram LMs, e.g., short-distance dependencies, and provide a flexible way to customize an LM by switching the underlying $n$-gram model without changing the neural model.

\subsection{Learning Objective\label{sec:objective}}
Ideally, to train a neural LM that approximates the residual function, one way is to re-define the $P_{NU}(x_k|\cdot)$ in Eq. (\ref{eq:neural_prob}) as follows:
\begin{align*}
P_{NU}(x_k|\seq{x}_{1}^{k-1}; \theta) = \mathcal{F}(&\seq{x}_{1}^{k-1})[x_k] + \nonumber \\ &\ \ P_{NG}(x_k|\seq{x}_{k-n+1}^{k-1}),
\end{align*}
where $\mathcal{F}(\cdot)$ is parameterized by the neural model $\theta$, and $P_{NG}(x_k|\cdot)$ is defined in Eq. (\ref{eq:kn}). Then we can optimize the MLE loss in Eq. (\ref{eq:mle}) based on the new $P_{NU}(x_k|\cdot)$, which is equivalent to approximate real-data distribution $\mathcal{G}$ by $\mathcal{F} + \mathcal{Q}$. However, directly optimizing this objective may have some problems. If $\mathcal{F}(\cdot)$ is unbounded, $P_{NU}$ defined in this equation may not be guaranteed as a valid probabilistic distribution. In contrast, if $\mathcal{F}(\cdot)$ is bounded as a valid distribution, this objective would become the ensemble of a neural LM and $n$-gram LM. Since $n$-gram is a weaker model, the ensemble of them is more likely to achieve worse performance than the vanilla neural LM, as shown in the experimental results of section \ref{sec:exp_lm}.

To address these issues, we propose to define residual approximation at the logits level. In the language modeling task, we can map the probabilistic distribution back to its logits and conduct residual learning as follows:
\begin{align}
\mathcal{F}^{\prime}(\seq{x}_{1}^{k-1}) :&= \mathrm{softmax}^{-1}\big(\mathcal{G}(\seq{x}_{1}^{k-1})\big) - \nonumber \\
&\ \ \ \ \ \ \ \ \mathrm{softmax}^{-1}\big(\mathcal{Q}(\seq{x}_{1}^{k-1})\big) \\
\mathrm{softmax}^{-1}(\seq{p}) &= \log \seq{p} + C,
\label{eq:rsm}\end{align}
where $\mathcal{F}^{\prime}(\cdot)$ is the residual function at the logits level,  $\mathrm{softmax}^{-1}(\seq{p})$ is the reverse function of $\mathrm{softmax}$ that maps the probabilistic distribution $\seq{p}$ to its logits, and $C$ is a constant. One reason that we conduct residual learning at the logits level is that logits are highly correlated to the final distribution. Moreover, since the value of logits is in the real number space, training the neural LM becomes more tractable by making sure that its logits are close to $\mathcal{F}^{\prime}(\seq{x}_{1}^{k-1})$. Therefore, the final $P_{NU}(x_k|\cdot)$ defined in our work is:
\begin{align}
    P_{NU}(x_k|\seq{x}_{1}^{k-1}; \theta) &
    = \mathrm{softmax}\Big(\mathcal{F}^{\prime}(\seq{x}_{1}^{k-1}) + \alpha\times \nonumber \\
    &\qquad\mathrm{softmax}^{-1}\big(\mathcal{Q}(\seq{x}_{1}^{k-1})\big)\Big)[x_k]
\label{eq:our_prob}\end{align}
where $\alpha$ is a hyper-parameter to control the smoothness of the logits of the $n$-gram distribution $\mathcal{Q}(\seq{x}_{1}^{k-1})$, and $\mathcal{F}^{\prime}(\cdot)$ is approximated by the logits $\phi(\seq{h}_k)$ of a neural LM. We can use the definition in Eq. (\ref{eq:our_prob}) to optimize the MLE loss in Eq. (\ref{eq:mle}).

\subsection{Relation to Re-weighting}
To better understand our approach, we can dive into the details of Eq. (\ref{eq:our_prob}). For simplicity, let us omit the condition $\seq{x}_{1}^{k-1}$ in this section:
\begin{align}
 P_{NU}(x_k|\cdot) &= \mathrm{softmax}\Big(\phi(\seq{h}_k) + \alpha\times \nonumber\\
&\quad\quad\quad\quad\big(\log P_{NG}(X|\cdot) + C\big)\Big)[x_k] \label{eq:line1}\\
&= \frac{\big(\mathrm{e}^C\big)^\alpha \big(\mathrm{e}^{\log P_{NG}(x_k|\cdot)}\big)^\alpha \mathrm{e}^{\phi(\seq{h}_k)[x_k]}}{Z},\label{eq:line2}
\end{align}
We apply the Eq. (\ref{eq:ngram_dist}) and (\ref{eq:rsm}) to get the explicit form of logits of the $n$-gram LM in Eq.(\ref{eq:line1}), and the definition of $\phi(\seq{h}_k)$ is the same as that in Eq. (\ref{eq:neural_prob}). In Eq. (\ref{eq:line2}), we expand the $\mathrm{softmax}$ function, where $Z$ is the normalization term. The numerator of Eq. (\ref{eq:line2}) has three terms. The first term $(\mathrm{e}^C)^\alpha$ is a constant for all the logit values, which does not affect the distribution. The middle term $(\mathrm{e}^{\log P_{NG}(x_k|\cdot)})^\alpha$ actually equals to $P_{NG}(x_k|\cdot)^\alpha$, which makes it be like the weight of the the logits of neural LM, i.e., the last term $\mathrm{e}^{\phi(\seq{h}_k)[x_k]}$ in Eq. (\ref{eq:line2}). When comparing with the vanilla neural LM, the golden-truth words are not equally important in the learning process of our approach. For golden-truth words that are well estimated by the $n$-gram LM, our approach would get high probabilities after $\mathrm{softmax}$, leading to a small loss value for the neural module. As a result, the neural model can spend more effort on difficult cases, such as predictions relying on long-distance dependencies, which are hard to be estimated by the $n$-gram LM.

\subsection{Discussion}

In this section, we propose a method to conduct residual learning between the neural and symbolic models, i.e., neural LM and $n$-gram LM. One of our expectations about the joint neuro-symbolic system is its better understanding of language. To evaluate this hypothesis, we can test our approach on standard language tasks, such as language modeling, machine translation, and summarization. The other expectation is the plug-and-lay property of our approach. For instance, if the testing data come from different domains, we can change the $\mathcal{Q}$ in Eq. (\ref{eq:our_prob}) by simply switching the used $n$-gram model.

\begin{table*}
\begin{center}
\resizebox{1.8\columnwidth}{!}{
    \begin{tabular}{c|l|ccccc|c}\toprule
    \# & \textbf{} &  \textbf{IT} & \textbf{Koran} & \textbf{Law}  & \textbf{Medical}  &\textbf{Subtitles}  & \textbf{AVG.} \\\hline
    1 & \textsc{\#Sent} & 222,927 & 17,982 & 467,309 & 248,099 & 500,000 & -- \\
    2 & \textsc{\#Word} & 2,585,965 & 4,512,266 & 15,348,052 & 4,512,266 & 5,125,239 & -- \\\midrule\midrule
    3 & \textsc{KenLM-5gram} & 95.89 & 35.51 & 15.74 & 24.00 & 101.99 & 54.63 \\
    4 & \textsc{GPT-2} & 66.49 & 35.34 & 9.93 & 15.18 & 77.34 & 40.86 \\
    5 & \ \ + \textsc{Finetune} & 53.69 & 26.77 & 9.43 & 12.96 &	69.33 & 34.44 \\
    6 & \ \ + \textsc{NgramRes} & 54.29 & 28.08 & 8.93 & 13.29 & 71.80 & 35.28 \\
    \bottomrule
    \end{tabular}
}
\end{center}
\caption{\label{tab:domain} Test perplexity of five domains. Results in lines 1-2 are the statistical information of each domain. Results in lines 3-6 are the perplexity scores of different approaches when testing on the five domains. The \textsc{GPT-2} and \textsc{NgramRes} (Line 4 and 6) approaches only train unified models for five domains, while the \textsc{Finetune} method (Line 5) trains a domain-specific model for each domain.}
\end{table*}

\begin{table}
\resizebox{1.0\columnwidth}{!}{
    \begin{tabular}{l|l|cc}\toprule
   \# & \textbf{Model} &  \textbf{\#Param} & \textbf{PPL} \\\hline
   1 & \cite{grave2016improving} - \textsc{LSTM} & -- & 40.8 \\
   2 & \cite{dauphin2017language} - \textsc{GCNN-8} & 229M & 37.2 \\
   3 & \cite{merity2018analysis} - \textsc{QRNN} & 151M & 33.0 \\
   4 & \cite{rae2018fast} - \textsc{Hebbian + Cache} & -- & 29.2 \\
   5 & \cite{baevski2018adaptive} - ADP & 247M & 18.7 \\\hline
   6 & \textsc{KenLM-5gram} &  -- & 116.4 \\
   7 & \textsc{ADP-Fairseq} &  247M & 18.9\\
   8 & \ \ \textsc{+ NgramRes} &  247M & 18.2 \\
   9 & \textsc{GPT-2 (BPE)} &  185M & 22.2\\
   10 & \ \ \textsc{+ Prob-Inter} &  185M & 60.2\\
   11 & \ \ \textsc{+ NgramRes} &  185M & 21.3 \\
    \bottomrule
    \end{tabular}
}
    \caption{\label{tab:general} Test perplexity on wikitext-103. Results in lines 1-5 are reported in previous works, and results in lines 6-11 are run by us. The \textsc{NgramRes} is our approach discussed in section \ref{sec:approach}.}
\end{table}

\section{Experiments}

In our work, we consider three kinds of natural language generation tasks: language modeling, machine translation, and summarization. For the language modeling task, we first evaluate the performance of our approach on the standard setting of the language modeling task. Then we turn to a domain adaptation setting.

\subsection{Language Modeling\label{sec:exp_lm}}

\paragraph{Setup} We use the wikitext-103 benchmark\footnote{Dataset provided by fairseq: \url{https://s3.amazonaws.com/research.metamind.io/wikitext/wikitext-103-v1.zip}} to evaluate the performance of our approach in the standard setting. The training set contains around 101M tokens. Following \citet{merity2016pointer}, tokens with a frequency lower than 3 have been replaced by the special token \texttt{<unk>} in the training datasets, and the number of remaining unique words is around 260k. For wikitext-103, we will train models at both word and subword levels. The subword-level data is preprocessed using \texttt{subword-nmt}\footnote{\url{https://github.com/rsennrich/subword-nmt}} \cite{sennrich-etal-2016-neural}, where the number of merge operation is set to 32k.

We use \texttt{fairseq}\footnote{\url{https://github.com/facebookresearch/fairseq}} \cite{ott2019fairseq} as the code base of our neural modules. We implement our approach on two popular neural language models, GPT-2 base \cite{radford2019language} and Adaptive Input (ADP) \cite{baevski2018adaptive}. For the ADP model, we follow the original hyper-parameters and use the code released by \citet{baevski2018adaptive} in \texttt{fairseq}\footnote{\url{https://github.com/facebookresearch/fairseq/blob/main/examples/language_model/README.adaptive_inputs.md}} to train the model on word-level data. Since the vocabulary size of the word-level data is too large, we train the GPT-2 base model on the subword-level data. For those neural models, we mostly use their default hyper-parameters reported in their paper \cite{baevski2018adaptive, radford2019language} and train those models from random initialization. Regarding to the $n$-gram model, we use the \texttt{KenLM}\footnote{\url{https://github.com/kpu/kenlm}} \cite{heafield-2011-kenlm} to train $n$-gram models on both the word-level and subword-level data of wikitext-103. The $n$ is set to 5 in our work. To make the perplexities of different models comparable, we report all the perplexity scores at the word level. For subword-level data, the word-level probability is the product of its subword tokens, following \citet{baevski2018adaptive}.

When training our approach \textsc{NgramRes}, we will hybrid the \textsc{KenLM-5gram} model and the neural model, i.e., \textsc{GPT-2} and \textsc{ADP}, using the residual learning method discussed in section \ref{sec:approach}. The hyper-parameter $\alpha$ in Eq. (\ref{eq:our_prob}) is tuned according to the performance on the validation dataset.

\paragraph{Results} 

As shown in Table \ref{tab:general}, we evaluate our approach on the wikitext-103 benchmark. Although the macro performance of \textsc{KenLM-5gram} (Line 6) on the test set is poor, it is still able to promote the performance of our approach. When comparing our approach (Line 8 and 11) with the vanilla neural models (Line 7 and 9), our approach steadily outperforms \textsc{ADP-Fairseq}\footnote{This is the result by running the officially released code of ADP} and \textsc{GPT-2} by 0.7 and 0.9 PPL scores, respectively. According to these results, \textsc{NgramRes} is able to improve the model performance without changing the architecture and the number of parameters.

Moreover, we also compare our method with a straightforward baseline \textsc{Prob-Inter}, as discussed in section \ref{sec:approach}. The \textsc{Prob-Inter} baseline directly interpolates the probabilistic distribution of \textsc{KenLM-5gram} and \textsc{GPT-2}. The performance of \textsc{Prob-Inter} is better than the \textsc{KenLM-5gram} but worse than the vanilla \textsc{GPT-2}, making it like the ensemble of the two models, as we discussed in the section \ref{sec:approach}.

\begin{table*}[t]
\begin{center}
\resizebox{1.8\columnwidth}{!}{
    \begin{tabular}{l|cccc|c}\toprule
    \textbf{Model} &  \textbf{En} $\Rightarrow$ \textbf{Fr} & \textbf{En} $\Rightarrow$ \textbf{Es} & \textbf{En} $\Rightarrow$ \textbf{Vi} &  \textbf{En} $\Rightarrow$ \textbf{De} & \textbf{AVG.} \\\midrule
    \textsc{Transformer} & 39.96 & 36.99 & 28.55 & 27.79 & 33.32 \\
    \ \ + \textsc{NgramRes} & 40.27 & 37.27 & 29.60 & 28.05 & 33.79 \\
    \ \ + \textsc{NgramRes-Anneal} & 40.49 & 37.07 & 29.92 & 28.41 & 33.97 \\
    \bottomrule
    \end{tabular}
}
\end{center}
\caption{\label{tab:mt} BLEU scores on IWSLT. The \textsc{Transformer} model is the baseline, and \textsc{NgramRes} and \textsc{NgramRes-Anneal} are two variants of our approach. Comparing with \textsc{NgramRes}, the \textsc{NgramRes-Anneal} decreases the value of $\alpha$ in Eq. (\ref{eq:our_prob}) linearly in the first 10k steps of model training.}
\end{table*}

\begin{table*}[ht]
\centering
\label{table:cnndm}
\resizebox{1.8\columnwidth}{!}{
\begin{tabular}{l|cccc}
\toprule
\textbf{Model} & \textbf{ROUGE-1} & \textbf{ROUGE-2} & \textbf{ROUGE-L} \\
\midrule
Pointer-generator + Coverage~\cite{see2017get} & 39.53 & 17.28 & 36.38 \\
Mask Attention Network~\cite{fan2021mask} & 40.98 & 18.29 & 37.88 \\
BertSum~\cite{liu2019text} & 42.13 & 19.60 & 39.18 \\
UniLM~\cite{dong2019unified} & 43.08 & 20.43 & 40.34 \\
UniLM V2~\cite{bao2020unilmv2} & 43.16 & 20.42 & 40.14 \\
ERNIE-GEN-large~\cite{xiao2021ernie} & 44.02 & 21.17 & 41.26 \\
PEGASUS~\cite{zhang2020pegasus} & 44.17 & 21.47 & 41.11 \\
ProphetNet~\cite{qi2020prophetnet} & 44.20 & 21.17 & 41.30 \\
PALM~\cite{bi2020palm} & 44.30 & 21.12 & 41.14 \\
\midrule
\textsc{BART-large}~\cite{lewis-etal-2020-bart} & 44.11 & 21.21 & 40.83\\
\ \ + \textsc{NgramRes} & 44.41 & 21.36 & 41.19 \\
\bottomrule
\end{tabular}
}
\caption{ROUGE scores on the test set of CNN/DailyMail dataset.\label{tab:sum}}
\end{table*}

\subsection{Language Modeling: Multi-Domain}

In this setting, we will evaluate the performance of adapting our approach to a specific domain by changing the used $n$-gram model.

\paragraph{Setup} In the multi-domain setting, we use the English side of a bilingual dataset with 5 domains \cite{aharoni-goldberg-2020-unsupervised}, i.e., IT, Koran, Law, Medical, and Subtitles. The statistical information of this dataset is shown in Table \ref{tab:domain}.  we apply \texttt{subword-nmt} on the joint training data of five domains, and the number of the merge operation is also 32k.

Following the standard setting of the language modeling task, we use \textsc{GPT-2} base \cite{radford2019language} as the neural model. We train and select \textsc{GPT-2} model on the mixed data from five domains, and report the word-level perplexity on the test data of each domain independently. The \textsc{GPT-2 + Finetune} method will adapt the parameters of \textsc{GPT-2} model on the corresponding domain before testing. For our approach \textsc{NgramRes}, we train a 5-gram LM for each specific domain and switch the used 5-gram model to the corresponding domain during training and testing. It is worth noting that the neural parameters of \textsc{NgramRes} are fixed when testing.

\paragraph{Results} 
The experimental results are shown in Table \ref{tab:domain}. For \textsc{GPT-2} and \textsc{NgramRes} (Line 4 and 6), we train unified neural models on mixed data of five domains and evaluate their performances on the test data of five domains one by one. Results show that our approach can outperform the vanilla neural model \textsc{GPT-2} by a large margin. Since the \textsc{NgramRes} approach stores a lot of domain-specific information in the 5-gram LM, we hypothesize that the neural module is able to learn useful and complementary knowledge during training, leading to the performance gain.

In the line of \textsc{+ Finetune}, we also report the results of fine-tuning the \textsc{GPT-2} model on each testing domain. It surprised us that the performances of our approach are very close to those of the \textsc{Finetune} method. The \textsc{NgramRes} even outperforms \textsc{Finetune} slightly on the Law domain. Moreover, compared with the \textsc{Finetune}, one advantage of our approach is its low cost of adapting our model to the testing domain, since we only need to replace the used $5$-gram model in a plug-and-play manner.

\subsection{Machine Translation}

Next, we evaluate our approach on a popular sequence-to-sequence task, namely, machine translation. Note that we only integrate our approach into the decoder side of the encoder-decoder model. 

\paragraph{Setup} We conduct the experiments of machine translation on IWSLT14 (En $\Rightarrow$ Fr, Es, De) and IWSLT15 (En $\Rightarrow$ Vi). The IWSLT14 datasets\footnote{\url{https://wit3.fbk.eu/2014-01}} of three language pairs are preprocessed following the script provided by fairseq\footnote{\url{https://github.com/facebookresearch/fairseq/blob/main/examples/translation/prepare-iwslt14.sh}}, where the evaluation data is sampled from the whole dataset and the test data is the concatenation of \textit{dev2011}, \textit{tst2012}, \textit{tst2012}. There is no overlap between train, validation, and test sets. For IWSLT15, we use the train, evaluation, and test data preprocessed and released by Stanford\footnote{\url{https://nlp.stanford.edu/projects/nmt/}} \cite{Luong-Manning:iwslt15}. The results are reported using tokenized SacreBLEU\footnote{\url{https://github.com/mjpost/sacrebleu}} \cite{post-2018-call}.

We use \texttt{fairseq} as our code base. We use the Transformer model as our architecture\footnote{The used architecture code in \texttt{fairseq} is \texttt{transformer\_iwslt\_de\_en}} for all the translation models. The Transformer model has 6 encoder layers and 6 decoder layers. Since the IWSLT datasets are small,  the hidden size of FFN sublayers is set to 1024, the number of attention heads is set to 4, the dropout rate is set to 0.3, and the weight decay rate is set to 0.001. We set other hyper-parameters according to the default setting of \citet{vaswani2017attention}. All the translation models are trained for 30 epochs from random initialization.

The implementation details of the $n$-gram model and our approach are similar to that in the language modeling task. For the translation task, we only use the target data, i.e., the X side of En$\Rightarrow$X data, to train the \textsc{KenLM-5gram} LM.

\paragraph{Results} The results of machine translation are shown in Table \ref{tab:mt}. We implement two variants of our approaches, namely, \textsc{NgramRes} and \textsc{NgramRes-Anneal}. The system of \textsc{NgramRes} only uses the 5-gram information on the decoder side, as we discussed in section \ref{sec:approach}. The difference between \textsc{NgramRes} and \textsc{NgramRes-Anneal} system is that the latter decreases the value of $\alpha$ linearly after each update . The alpha value becomes zero after 10k steps.

We find that both the two variants of our approaches outperform the \textsc{Transformer} model. The \textsc{NgramRes-Anneal} achieves the best results on each language pair, which means that the $n$-gram model is more critical for the beginning phase and may hurt the translation performance after that phase. According to \citet{voita-etal-2021-language}, the training of neural machine translation (NMT) systems undergoes three stages: target-side language modeling, learning the word-by-word translation, and learning to reorder. Therefore, we hypothesize that the use of the $n$-gram model in the whole training procedure may over-emphasize the importance of target-side language modeling in NMT, having a negative impact on the next two stages.

\subsection{Abstractive Summarization}
Lastly, we evaluate our approach on another popular sequence-to-sequence task, namely, abstractive summarization. Like machine translation, our approach is applied to the decoder side of the encoder-decoder model.

\paragraph{Setup} For the abstractive summarization task, we preprocess the CNN/DailyMail dataset following the script provided by \texttt{fairseq}\footnote{\url{https://github.com/facebookresearch/fairseq/blob/main/examples/bart/README.summarization.md}}. The evaluation metrics of the summarization task are ROUGE scores, i.e., ROUGE-1, ROUGE-2, and ROUGE-L \cite{lin-2004-rouge}\footnote{\url{https://github.com/pltrdy/files2rouge}}.

We follow the setting of previous works and finetune the pre-trained \textsc{BART-large} model \cite{lewis-etal-2020-bart} on the CNN/DailyMail dataset for 20k updates. We train the \textsc{KenLM-5gram} LM on the joint data of its source and summarization text.

\paragraph{Results} The summarization task is also a sequence-to-sequence task, where the source text and summarization are in the same language and share similar semantics. As shown in Table \ref{tab:sum}, in this task, our approach is still able to improve the performance of the strong baseline model \textsc{BART-large}, without any change in the model architecture.

Different from the machine translation task, we find that using a fixed $\alpha$ value achieves better performance than annealing it. The reason may be that the target-side language modeling plays a more important role in the summarization task because summarization is more like monolingual text generation in a constrained context.

\section{Conclusion and Future Work}

This work aims to learn a neural LM that approximates the information that has not been captured by $n$-gram LM. To achieve this goal, we propose a residual learning approach to force the two neural and symbolic models, i.e., the neural LM and $n$-gram LM, to learn complementary information. We conduct extensive experiments to evaluate the performance of the proposed approach. In our experiments, we find that our neuro-symbolic system can not only improve the performance of recent state-of-the-art neural models consistently and considerable on three typical language tasks (including language modeling, machine translation, and summarization) but also exhibits a good plug-and-play property on the multi-domain language modeling task.

The $n$-gram LM has lots of attractive properties that we have not explored in this work. First, the $n$-gram model has good interpretability. The behavior of $n$-gram LM is easier to understand than the weights of neurons from the perspective of humans. In the future, we want to leverage the property of the $n$-gram model to better understand the decision-making process of the neural LM.
Second, controlling the system predictions through the $n$-gram model may have a big potential. As observed in our multi-domain experiments, we are able to customize an LM by switching the underlying $n$-gram model without changing the neural part. It is also interesting to explore how to control the model output at a fine-grained level using the $n$-gram LM.

\section*{Limitations}

We believe there are two limitations in our approach. First, since the estimation of the prediction distribution of $n$-gram models relies on CPU, the estimation speed by $n$-gram models may be slow when using a big batch size ($>> 8192 * 8$). Second, the performance gain of our current approach on high-resource datasets is not big. For instance, we also evaluate the performance of \textsc{Transformer + NgramRes} on WMT14 En-De \cite{vaswani2017attention}, but the improvement is only 0.15 BLEU score. These limitations urge us to propose more efficient and effective approaches in future works.

\section*{Acknowledgement}

We are particularly grateful for the help from Xiaojiang Liu, because this project would never have been conceived and completed without his generous and selfless support. We also want to thank the insightful discussions with Yixuan Su and the valuable comments from our anonymous reviewers, area chairs, and senior area chairs.

\bibliography{anthology,custom}

\begin{thebibliography}{50}
\expandafter\ifx\csname natexlab\endcsname\relax\def\natexlab#1{#1}\fi

\bibitem[{Aharoni and Goldberg(2020)}]{aharoni-goldberg-2020-unsupervised}
Roee Aharoni and Yoav Goldberg. 2020.
\newblock \href {https://doi.org/10.18653/v1/2020.acl-main.692} {Unsupervised
  domain clusters in pretrained language models}.
\newblock In \emph{Proceedings of the 58th Annual Meeting of the Association
  for Computational Linguistics}, pages 7747--7763, Online. Association for
  Computational Linguistics.

\bibitem[{Baevski and Auli(2019)}]{baevski2018adaptive}
Alexei Baevski and Michael Auli. 2019.
\newblock \href {https://openreview.net/forum?id=ByxZX20qFQ} {Adaptive input
  representations for neural language modeling}.
\newblock In \emph{7th International Conference on Learning Representations,
  {ICLR} 2019, New Orleans, LA, USA, May 6-9, 2019}. OpenReview.net.

\bibitem[{Bahdanau et~al.(2015)Bahdanau, Cho, and Bengio}]{bahdanau2015neural}
Dzmitry Bahdanau, Kyunghyun Cho, and Yoshua Bengio. 2015.
\newblock \href {http://arxiv.org/abs/1409.0473} {Neural machine translation by
  jointly learning to align and translate}.
\newblock In \emph{3rd International Conference on Learning Representations,
  {ICLR} 2015, San Diego, CA, USA, May 7-9, 2015, Conference Track
  Proceedings}.

\bibitem[{Bahl et~al.(1983)Bahl, Jelinek, and Mercer}]{bahl1983maximum}
Lalit~R. Bahl, Frederick Jelinek, and Robert~L. Mercer. 1983.
\newblock \href {https://doi.org/10.1109/TPAMI.1983.4767370} {A maximum
  likelihood approach to continuous speech recognition}.
\newblock \emph{{IEEE} Trans. Pattern Anal. Mach. Intell.}, 5(2):179--190.

\bibitem[{Bao et~al.(2020)Bao, Dong, Wei, Wang, Yang, Liu, Wang, Gao, Piao,
  Zhou et~al.}]{bao2020unilmv2}
Hangbo Bao, Li~Dong, Furu Wei, Wenhui Wang, Nan Yang, Xiaodong Liu, Yu~Wang,
  Jianfeng Gao, Songhao Piao, Ming Zhou, et~al. 2020.
\newblock Unilmv2: Pseudo-masked language models for unified language model
  pre-training.
\newblock In \emph{International Conference on Machine Learning}, pages
  642--652. PMLR.

\bibitem[{Bi et~al.(2020)Bi, Li, Wu, Yan, Wang, Huang, Huang, and
  Si}]{bi2020palm}
Bin Bi, Chenliang Li, Chen Wu, Ming Yan, Wei Wang, Songfang Huang, Fei Huang,
  and Luo Si. 2020.
\newblock Palm: Pre-training an autoencoding\&autoregressive language model for
  context-conditioned generation.
\newblock In \emph{Proceedings of the 2020 Conference on Empirical Methods in
  Natural Language Processing (EMNLP)}, pages 8681--8691.

\bibitem[{Brown et~al.(1990)Brown, Cocke, Pietra, Pietra, Jelinek, Lafferty,
  Mercer, and Roossin}]{brown1990statistical}
Peter~F. Brown, John Cocke, Stephen~Della Pietra, Vincent J.~Della Pietra,
  Frederick Jelinek, John~D. Lafferty, Robert~L. Mercer, and Paul~S. Roossin.
  1990.
\newblock A statistical approach to machine translation.
\newblock \emph{Comput. Linguistics}, 16(2):79--85.

\bibitem[{Cai et~al.(2021)Cai, Wang, Li, Lam, and Liu}]{cai2021neural}
Deng Cai, Yan Wang, Huayang Li, Wai Lam, and Lemao Liu. 2021.
\newblock Neural machine translation with monolingual translation memory.
\newblock In \emph{Proceedings of the 59th Annual Meeting of the Association
  for Computational Linguistics and the 11th International Joint Conference on
  Natural Language Processing (Volume 1: Long Papers)}, pages 7307--7318.

\bibitem[{Cai et~al.(2022)Cai, Wang, Liu, and Shi}]{cai2022recent}
Deng Cai, Yan Wang, Lemao Liu, and Shuming Shi. 2022.
\newblock Recent advances in retrieval-augmented text generation.
\newblock In \emph{Proceedings of the 45th International ACM SIGIR Conference
  on Research and Development in Information Retrieval}, pages 3417--3419.

\bibitem[{Chen et~al.(2019)Chen, Suresh, Mathews, Wong, Allauzen, Beaufays, and
  Riley}]{chen-etal-2019-federated}
Mingqing Chen, Ananda~Theertha Suresh, Rajiv Mathews, Adeline Wong, Cyril
  Allauzen, Fran{\c{c}}oise Beaufays, and Michael Riley. 2019.
\newblock \href {https://doi.org/10.18653/v1/K19-1012} {Federated learning of
  n-gram language models}.
\newblock In \emph{Proceedings of the 23rd Conference on Computational Natural
  Language Learning (CoNLL)}, pages 121--130, Hong Kong, China. Association for
  Computational Linguistics.

\bibitem[{Chen and Goodman(1996)}]{chen1996empirical}
Stanley~F. Chen and Joshua Goodman. 1996.
\newblock \href {https://doi.org/10.3115/981863.981904} {An empirical study of
  smoothing techniques for language modeling}.
\newblock In \emph{34th Annual Meeting of the Association for Computational
  Linguistics, 24-27 June 1996, University of California, Santa Cruz,
  California, USA, Proceedings}, pages 310--318. Morgan Kaufmann Publishers /
  {ACL}.

\bibitem[{Chung et~al.(2014)Chung, G{\"{u}}l{\c{c}}ehre, Cho, and
  Bengio}]{chung2014emperical}
Junyoung Chung, {\c{C}}aglar G{\"{u}}l{\c{c}}ehre, KyungHyun Cho, and Yoshua
  Bengio. 2014.
\newblock \href {http://arxiv.org/abs/1412.3555} {Empirical evaluation of gated
  recurrent neural networks on sequence modeling}.
\newblock \emph{CoRR}, abs/1412.3555.

\bibitem[{Dauphin et~al.(2017)Dauphin, Fan, Auli, and
  Grangier}]{dauphin2017language}
Yann~N. Dauphin, Angela Fan, Michael Auli, and David Grangier. 2017.
\newblock \href {http://proceedings.mlr.press/v70/dauphin17a.html} {Language
  modeling with gated convolutional networks}.
\newblock In \emph{Proceedings of the 34th International Conference on Machine
  Learning, {ICML} 2017, Sydney, NSW, Australia, 6-11 August 2017}, volume~70
  of \emph{Proceedings of Machine Learning Research}, pages 933--941. {PMLR}.

\bibitem[{Dong et~al.(2019)Dong, Yang, Wang, Wei, Liu, Wang, Gao, Zhou, and
  Hon}]{dong2019unified}
Li~Dong, Nan Yang, Wenhui Wang, Furu Wei, Xiaodong Liu, Yu~Wang, Jianfeng Gao,
  Ming Zhou, and Hsiao-Wuen Hon. 2019.
\newblock Unified language model pre-training for natural language
  understanding and generation.
\newblock \emph{Advances in Neural Information Processing Systems}, 32.

\bibitem[{Fan et~al.(2021)Fan, Gong, Liu, Wei, Wang, Jiao, Duan, Zhang, and
  Huang}]{fan2021mask}
Zhihao Fan, Yeyun Gong, Dayiheng Liu, Zhongyu Wei, Siyuan Wang, Jian Jiao, Nan
  Duan, Ruofei Zhang, and Xuan-Jing Huang. 2021.
\newblock Mask attention networks: Rethinking and strengthen transformer.
\newblock In \emph{Proceedings of the 2021 Conference of the North American
  Chapter of the Association for Computational Linguistics: Human Language
  Technologies}, pages 1692--1701.

\bibitem[{Grave et~al.(2017)Grave, Joulin, and Usunier}]{grave2016improving}
Edouard Grave, Armand Joulin, and Nicolas Usunier. 2017.
\newblock \href {https://openreview.net/forum?id=B184E5qee} {Improving neural
  language models with a continuous cache}.
\newblock In \emph{5th International Conference on Learning Representations,
  {ICLR} 2017, Toulon, France, April 24-26, 2017, Conference Track
  Proceedings}. OpenReview.net.

\bibitem[{He et~al.(2016)He, Zhang, Ren, and Sun}]{he2016deep}
Kaiming He, Xiangyu Zhang, Shaoqing Ren, and Jian Sun. 2016.
\newblock \href {https://doi.org/10.1109/CVPR.2016.90} {Deep residual learning
  for image recognition}.
\newblock In \emph{2016 {IEEE} Conference on Computer Vision and Pattern
  Recognition, {CVPR} 2016, Las Vegas, NV, USA, June 27-30, 2016}, pages
  770--778. {IEEE} Computer Society.

\bibitem[{He et~al.(2021)He, Ravula, Kanagal, and Ainslie}]{he2021realFormer}
Ruining He, Anirudh Ravula, Bhargav Kanagal, and Joshua Ainslie. 2021.
\newblock \href {https://doi.org/10.18653/v1/2021.findings-acl.81} {Realformer:
  Transformer likes residual attention}.
\newblock In \emph{Findings of the Association for Computational Linguistics:
  {ACL/IJCNLP} 2021, Online Event, August 1-6, 2021}, volume {ACL/IJCNLP} 2021
  of \emph{Findings of {ACL}}, pages 929--943. Association for Computational
  Linguistics.

\bibitem[{Heafield(2011)}]{heafield-2011-kenlm}
Kenneth Heafield. 2011.
\newblock \href {https://www.aclweb.org/anthology/W11-2123} {{K}en{LM}: Faster
  and smaller language model queries}.
\newblock In \emph{Proceedings of the Sixth Workshop on Statistical Machine
  Translation}, pages 187--197, Edinburgh, Scotland. Association for
  Computational Linguistics.

\bibitem[{Heafield et~al.(2013)Heafield, Pouzyrevsky, Clark, and
  Koehn}]{heafield-etal-2013-scalable}
Kenneth Heafield, Ivan Pouzyrevsky, Jonathan~H. Clark, and Philipp Koehn. 2013.
\newblock \href {https://www.aclweb.org/anthology/P13-2121} {Scalable modified
  {K}neser-{N}ey language model estimation}.
\newblock In \emph{Proceedings of the 51st Annual Meeting of the Association
  for Computational Linguistics (Volume 2: Short Papers)}, pages 690--696,
  Sofia, Bulgaria. Association for Computational Linguistics.

\bibitem[{Hochreiter and Schmidhuber(1997)}]{hochreiter1997long}
Sepp Hochreiter and J{\"{u}}rgen Schmidhuber. 1997.
\newblock \href {https://doi.org/10.1162/neco.1997.9.8.1735} {Long short-term
  memory}.
\newblock \emph{Neural Comput.}, 9(8):1735--1780.

\bibitem[{Holtzman et~al.(2020)Holtzman, Buys, Du, Forbes, and
  Choi}]{DBLP:conf/iclr/HoltzmanBDFC20}
Ari Holtzman, Jan Buys, Li~Du, Maxwell Forbes, and Yejin Choi. 2020.
\newblock \href {https://openreview.net/forum?id=rygGQyrFvH} {The curious case
  of neural text degeneration}.
\newblock In \emph{8th International Conference on Learning Representations,
  {ICLR} 2020, Addis Ababa, Ethiopia, April 26-30, 2020}. OpenReview.net.

\bibitem[{Huang et~al.(2015)Huang, Zhang, Zhou, and Zong}]{huang2015input}
Guoping Huang, Jiajun Zhang, Yu~Zhou, and Chengqing Zong. 2015.
\newblock \href {http://ijcai.org/Abstract/15/168} {A new input method for
  human translators: Integrating machine translation effectively and
  imperceptibly}.
\newblock In \emph{Proceedings of the Twenty-Fourth International Joint
  Conference on Artificial Intelligence, {IJCAI} 2015, Buenos Aires, Argentina,
  July 25-31, 2015}, pages 1163--1169. {AAAI} Press.

\bibitem[{Jurafsky(2000)}]{jurafsky2000speech}
Dan Jurafsky. 2000.
\newblock \emph{Speech \& language processing}.
\newblock Pearson Education India.

\bibitem[{Kneser and Ney(1995)}]{kneser1995improved}
Reinhard Kneser and Hermann Ney. 1995.
\newblock \href {https://doi.org/10.1109/ICASSP.1995.479394} {Improved
  backing-off for m-gram language modeling}.
\newblock In \emph{1995 International Conference on Acoustics, Speech, and
  Signal Processing, {ICASSP} '95, Detroit, Michigan, USA, May 08-12, 1995},
  pages 181--184. {IEEE} Computer Society.

\bibitem[{Lewis et~al.(2020)Lewis, Liu, Goyal, Ghazvininejad, Mohamed, Levy,
  Stoyanov, and Zettlemoyer}]{lewis-etal-2020-bart}
Mike Lewis, Yinhan Liu, Naman Goyal, Marjan Ghazvininejad, Abdelrahman Mohamed,
  Omer Levy, Veselin Stoyanov, and Luke Zettlemoyer. 2020.
\newblock \href {https://doi.org/10.18653/v1/2020.acl-main.703} {{BART}:
  Denoising sequence-to-sequence pre-training for natural language generation,
  translation, and comprehension}.
\newblock In \emph{Proceedings of the 58th Annual Meeting of the Association
  for Computational Linguistics}, pages 7871--7880, Online. Association for
  Computational Linguistics.

\bibitem[{Li et~al.(2022)Li, Su, Cai, Wang, and Liu}]{li2022survey}
Huayang Li, Yixuan Su, Deng Cai, Yan Wang, and Lemao Liu. 2022.
\newblock A survey on retrieval-augmented text generation.
\newblock \emph{arXiv preprint arXiv:2202.01110}.

\bibitem[{Li et~al.(2017)Li, Lam, Bing, and Wang}]{li-etal-2017-deep}
Piji Li, Wai Lam, Lidong Bing, and Zihao Wang. 2017.
\newblock \href {https://doi.org/10.18653/v1/D17-1222} {Deep recurrent
  generative decoder for abstractive text summarization}.
\newblock In \emph{Proceedings of the 2017 Conference on Empirical Methods in
  Natural Language Processing}, pages 2091--2100, Copenhagen, Denmark.
  Association for Computational Linguistics.

\bibitem[{Lin(2004)}]{lin-2004-rouge}
Chin-Yew Lin. 2004.
\newblock \href {https://aclanthology.org/W04-1013} {{ROUGE}: A package for
  automatic evaluation of summaries}.
\newblock In \emph{Text Summarization Branches Out}, pages 74--81, Barcelona,
  Spain. Association for Computational Linguistics.

\bibitem[{Liu and Lapata(2019)}]{liu2019text}
Yang Liu and Mirella Lapata. 2019.
\newblock Text summarization with pretrained encoders.
\newblock In \emph{Proceedings of the 2019 Conference on Empirical Methods in
  Natural Language Processing and the 9th International Joint Conference on
  Natural Language Processing (EMNLP-IJCNLP)}, pages 3730--3740.

\bibitem[{Luong and Manning(2015)}]{Luong-Manning:iwslt15}
Minh-Thang Luong and Christopher~D. Manning. 2015.
\newblock Stanford neural machine translation systems for spoken language
  domain.
\newblock In \emph{International Workshop on Spoken Language Translation}, Da
  Nang, Vietnam.

\bibitem[{Merity et~al.(2018)Merity, Keskar, and Socher}]{merity2018analysis}
Stephen Merity, Nitish~Shirish Keskar, and Richard Socher. 2018.
\newblock \href {http://arxiv.org/abs/1803.08240} {An analysis of neural
  language modeling at multiple scales}.
\newblock \emph{CoRR}, abs/1803.08240.

\bibitem[{Merity et~al.(2017)Merity, Xiong, Bradbury, and
  Socher}]{merity2016pointer}
Stephen Merity, Caiming Xiong, James Bradbury, and Richard Socher. 2017.
\newblock \href {https://openreview.net/forum?id=Byj72udxe} {Pointer sentinel
  mixture models}.
\newblock In \emph{5th International Conference on Learning Representations,
  {ICLR} 2017, Toulon, France, April 24-26, 2017, Conference Track
  Proceedings}. OpenReview.net.

\bibitem[{Ott et~al.(2019)Ott, Edunov, Baevski, Fan, Gross, Ng, Grangier, and
  Auli}]{ott2019fairseq}
Myle Ott, Sergey Edunov, Alexei Baevski, Angela Fan, Sam Gross, Nathan Ng,
  David Grangier, and Michael Auli. 2019.
\newblock fairseq: A fast, extensible toolkit for sequence modeling.
\newblock In \emph{Proceedings of NAACL-HLT 2019: Demonstrations}.

\bibitem[{Post(2018)}]{post-2018-call}
Matt Post. 2018.
\newblock \href {https://www.aclweb.org/anthology/W18-6319} {A call for clarity
  in reporting {BLEU} scores}.
\newblock In \emph{Proceedings of the Third Conference on Machine Translation:
  Research Papers}, pages 186--191, Belgium, Brussels. Association for
  Computational Linguistics.

\bibitem[{Prakash et~al.(2016)Prakash, Hasan, Lee, Datla, Qadir, Liu, and
  Farri}]{prakash2016neural}
Aaditya Prakash, Sadid~A. Hasan, Kathy Lee, Vivek~V. Datla, Ashequl Qadir, Joey
  Liu, and Oladimeji Farri. 2016.
\newblock \href {https://aclanthology.org/C16-1275/} {Neural paraphrase
  generation with stacked residual {LSTM} networks}.
\newblock In \emph{{COLING} 2016, 26th International Conference on
  Computational Linguistics, Proceedings of the Conference: Technical Papers,
  December 11-16, 2016, Osaka, Japan}, pages 2923--2934. {ACL}.

\bibitem[{Qi et~al.(2020)Qi, Yan, Gong, Liu, Duan, Chen, Zhang, and
  Zhou}]{qi2020prophetnet}
Weizhen Qi, Yu~Yan, Yeyun Gong, Dayiheng Liu, Nan Duan, Jiusheng Chen, Ruofei
  Zhang, and Ming Zhou. 2020.
\newblock Prophetnet: Predicting future n-gram for
  sequence-to-sequencepre-training.
\newblock In \emph{Findings of the Association for Computational Linguistics:
  EMNLP 2020}, pages 2401--2410.

\bibitem[{Radford et~al.(2019)Radford, Wu, Child, Luan, Amodei, and
  Sutskever}]{radford2019language}
Alec Radford, Jeff Wu, Rewon Child, David Luan, Dario Amodei, and Ilya
  Sutskever. 2019.
\newblock Language models are unsupervised multitask learners.

\bibitem[{Rae et~al.(2018)Rae, Dyer, Dayan, and Lillicrap}]{rae2018fast}
Jack~W. Rae, Chris Dyer, Peter Dayan, and Timothy~P. Lillicrap. 2018.
\newblock \href {http://proceedings.mlr.press/v80/rae18a.html} {Fast parametric
  learning with activation memorization}.
\newblock In \emph{Proceedings of the 35th International Conference on Machine
  Learning, {ICML} 2018, Stockholmsm{\"{a}}ssan, Stockholm, Sweden, July 10-15,
  2018}, volume~80 of \emph{Proceedings of Machine Learning Research}, pages
  4225--4234. {PMLR}.

\bibitem[{See et~al.(2017)See, Liu, and Manning}]{see2017get}
Abigail See, Peter~J Liu, and Christopher~D Manning. 2017.
\newblock Get to the point: Summarization with pointer-generator networks.
\newblock In \emph{Proceedings of the 55th Annual Meeting of the Association
  for Computational Linguistics (Volume 1: Long Papers)}, pages 1073--1083.

\bibitem[{Sennrich et~al.(2016)Sennrich, Haddow, and
  Birch}]{sennrich-etal-2016-neural}
Rico Sennrich, Barry Haddow, and Alexandra Birch. 2016.
\newblock \href {https://doi.org/10.18653/v1/P16-1162} {Neural machine
  translation of rare words with subword units}.
\newblock In \emph{Proceedings of the 54th Annual Meeting of the Association
  for Computational Linguistics (Volume 1: Long Papers)}, pages 1715--1725,
  Berlin, Germany. Association for Computational Linguistics.

\bibitem[{Su et~al.(2022)Su, Lan, Wang, Yogatama, Kong, and
  Collier}]{DBLP:journals/corr/abs-2202-06417}
Yixuan Su, Tian Lan, Yan Wang, Dani Yogatama, Lingpeng Kong, and Nigel Collier.
  2022.
\newblock A contrastive framework for neural text generation.
\newblock \emph{arXiv preprint arXiv:2202.06417}.

\bibitem[{Vaswani et~al.(2017)Vaswani, Shazeer, Parmar, Uszkoreit, Jones,
  Gomez, Kaiser, and Polosukhin}]{vaswani2017attention}
Ashish Vaswani, Noam Shazeer, Niki Parmar, Jakob Uszkoreit, Llion Jones,
  Aidan~N. Gomez, Lukasz Kaiser, and Illia Polosukhin. 2017.
\newblock \href
  {https://proceedings.neurips.cc/paper/2017/hash/3f5ee243547dee91fbd053c1c4a845aa-Abstract.html}
  {Attention is all you need}.
\newblock In \emph{Advances in Neural Information Processing Systems 30: Annual
  Conference on Neural Information Processing Systems 2017, December 4-9, 2017,
  Long Beach, CA, {USA}}, pages 5998--6008.

\bibitem[{Voita et~al.(2021)Voita, Sennrich, and
  Titov}]{voita-etal-2021-language}
Elena Voita, Rico Sennrich, and Ivan Titov. 2021.
\newblock \href {https://doi.org/10.18653/v1/2021.emnlp-main.667} {Language
  modeling, lexical translation, reordering: The training process of {NMT}
  through the lens of classical {SMT}}.
\newblock In \emph{Proceedings of the 2021 Conference on Empirical Methods in
  Natural Language Processing}, pages 8478--8491, Online and Punta Cana,
  Dominican Republic. Association for Computational Linguistics.

\bibitem[{Wang and Tian(2016)}]{wang2016recurrent}
Yiren Wang and Fei Tian. 2016.
\newblock \href {https://doi.org/10.18653/v1/d16-1093} {Recurrent residual
  learning for sequence classification}.
\newblock In \emph{Proceedings of the 2016 Conference on Empirical Methods in
  Natural Language Processing, {EMNLP} 2016, Austin, Texas, USA, November 1-4,
  2016}, pages 938--943. The Association for Computational Linguistics.

\bibitem[{Welleck et~al.(2020)Welleck, Kulikov, Roller, Dinan, Cho, and
  Weston}]{DBLP:conf/iclr/WelleckKRDCW20}
Sean Welleck, Ilia Kulikov, Stephen Roller, Emily Dinan, Kyunghyun Cho, and
  Jason Weston. 2020.
\newblock \href {https://openreview.net/forum?id=SJeYe0NtvH} {Neural text
  generation with unlikelihood training}.
\newblock In \emph{8th International Conference on Learning Representations,
  {ICLR} 2020, Addis Ababa, Ethiopia, April 26-30, 2020}. OpenReview.net.

\bibitem[{Werlen et~al.(2018)Werlen, Pappas, Ram, and
  Popescu{-}Belis}]{werlen2018self}
Lesly~Miculicich Werlen, Nikolaos Pappas, Dhananjay Ram, and Andrei
  Popescu{-}Belis. 2018.
\newblock \href {https://doi.org/10.18653/v1/n18-1124} {Self-attentive residual
  decoder for neural machine translation}.
\newblock In \emph{Proceedings of the 2018 Conference of the North American
  Chapter of the Association for Computational Linguistics: Human Language
  Technologies, {NAACL-HLT} 2018, New Orleans, Louisiana, USA, June 1-6, 2018,
  Volume 1 (Long Papers)}, pages 1366--1379. Association for Computational
  Linguistics.

\bibitem[{Xiao et~al.(2021)Xiao, Zhang, Li, Sun, Tian, Wu, and
  Wang}]{xiao2021ernie}
Dongling Xiao, Han Zhang, Yukun Li, Yu~Sun, Hao Tian, Hua Wu, and Haifeng Wang.
  2021.
\newblock Ernie-gen: an enhanced multi-flow pre-training and fine-tuning
  framework for natural language generation.
\newblock In \emph{Proceedings of the Twenty-Ninth International Conference on
  International Joint Conferences on Artificial Intelligence}, pages
  3997--4003.

\bibitem[{Xu et~al.(2022)Xu, Liu, Yan, Cai, Li, and
  Li}]{DBLP:journals/corr/abs-2206-02369}
Jin Xu, Xiaojiang Liu, Jianhao Yan, Deng Cai, Huayang Li, and Jian Li. 2022.
\newblock \href {https://doi.org/10.48550/arXiv.2206.02369} {Learning to break
  the loop: Analyzing and mitigating repetitions for neural text generation}.
\newblock \emph{CoRR}, abs/2206.02369.

\bibitem[{Zhang et~al.(2020)Zhang, Zhao, Saleh, and Liu}]{zhang2020pegasus}
Jingqing Zhang, Yao Zhao, Mohammad Saleh, and Peter Liu. 2020.
\newblock Pegasus: Pre-training with extracted gap-sentences for abstractive
  summarization.
\newblock In \emph{International Conference on Machine Learning}, pages
  11328--11339. PMLR.

\end{thebibliography}
\bibliographystyle{acl_natbib}

\end{document}